# A Pervasive Framework for Human Detection and Tracking

**Fesatidis Georgios, Bratsos Dimitrios, Kostas Kolomvatsos**

Department of Informatics and Telecommunications

University of Thessaly

Papasiopoulou 2-4, 35131, Lamia Greece

emails: {fgeorgios, brdimitrios, kostasks}@uth.gr

**Abstract**

The advent of the Edge Computing (EC) leads to a huge ecosystem where numerous nodes can interact with data collection devices located close to end users. Human detection and tracking can be realized at edge nodes that perform the surveillance of an area under consideration through the assistance of a set of sensors (e.g., cameras). Our target is to incorporate the discussed functionalities to embedded devices present at the edge keeping their size limited while increasing their processing capabilities. In this paper, we propose two models for human detection accompanied by algorithms for tracing the corresponding trajectories. We provide the description of the proposed models and extend them to meet the challenges of the problem. Our evaluation aims at identifying models' accuracy while presenting their requirements to have them executed in embedded devices.

## 1. Introduction

The Pervasive Computing (PC) paradigm targets to the placement of numerous devices around end users to support innovative applications. These devices can be connected with the network and become the points where data can be collected and transferred to the Edge and Cloud infrastructures for further processing. One can find multiple efforts that propose the use of monitoring mechanisms upon contextual streams for event detection at the Internet of Things (IoT) and Edge Computing (EC) [1], [15], [16]. A research question relates to the definition of efficient mechanisms that could be applied upon the contextual streams that will deliver the desired automated decision making. Affected by the application domain, various constraints should be supported. One significant constraint is the achievement of a limited latency in the provision of responses of processign activities. EC can offer the necessary infrastrucutre for performing various processing activities instead of relying on the Cloud and enjoy an increased latency [14]. We can easily discern an ecosystem of edge nodes connected with a number of end devices becoming the hosts of distributed datasets upon which the desired processing can be realized [2], [17]. EC nodes can convey the necessary intelligence to be adapted to their environment and support real time applications [13], [18]. One significant application is the detection of objects and more importantly the detection of humans based on video streams, i.e., the task of locating all the instances of human beings in an image (i.e., a video frame) [6]. Usually, a part of the image is compared with known human figures and deliver the probability of the detection. Video streams can be also adopted for tracking humans in a specific area and reason upon their trajectories. The application of human detection processing at the EC, we can conclude a platform that is able to have the relevant information and assist in various applications, e.g., evacuation planning, emergency management, pedestrian detection, etc.

Several researchers use Deep Learning (DL) models for human detection and classification purposes over traditional approaches, e.g., the perceptron model, Probabilistic Neural Networks (PNNs), Radial Basis Neural Networks (RBNs), Support Vector Machine (SVM), AdaBoost, etc [9]. The origin of such DL schemes are in the detection of objects in images, face recognition, image classification, speech recognition, text-to-speech generation, handwriting transcription and so on and so forth. Such schemes should be applied on the EC and the corresponding devices to have the discussed processing close to end users. In this paper, we propose a model for human detection and tracking and study the challegnes of its application in constrained environments. Our target is to continuously detect humans upon the feed of cameras placed at embedded devices (e.g., Coral TPUs, Raspberry Pis, etc) and efficiently support real time applications. The following list reports on the contributions of our paper: **(a)** We propose a combinations of models for human detection and tracking, i.e., we adopt the Single Shot Detector (SSD) [21] and the MobileNet [25] models enhanced to meet our requirements; **(b)** We setup an EC environment adopting Coral TPUs[1] for hosting and executing our model; **(c)** We perform an experimental evaluation upon the accuracy of the detection and the tracking process upon real video feeds.

The paper is organized as follows. Section 2 presents the prior work while Section 3 discusses the background information about the legacy models. Section 4 reports on the proposed models and Section 5 presents our experimental evaluation upon real video feeds. Finally, Section 6 discusses our conclusions and provides insights on our future research plans.

## 2. Related Work

There has been a rising interest around the field of object detection and tracking. Starting with the detectors, in general, they are separated in two main categories: (i) one-stage and (ii) two-stage. One-stage detectors produce their results in one step with representatives being are YOLO [4], SSD [21] and RetinaNet [20] with an anchor-free option also available using, but not limited to, CenterNet [7], CornerNet [19], FCOS [26]. Two stage models, initially, propose a set of ROIs (Regions Of Interest), then, a classifier processes these candidates. Representative two-stage detectors are R-CNN / fast R-CNN [10], faster R-CNN [24], R-FCN [5], and Libra R-CNN [23]. As for tracking, multiple algorithms are proposed like Kalman Filters [28], SORT [3], DeepSORT [27], CenterTrack [29] etc.

There is a wide of choices for the application domain involving object detection [22]. An important field is pedestrian detection, which is used in various applications such as surveillance footage, in order to detect any unwanted visitors or self-driving vehicles to be able to dodge people on the street. Another important field for this technology to be applied on is the face detection. Object detectors can be trained to be able to detect faces. Moreover, object detection is also used in highly controversial military applications and in the medical field to be able to detect tumors and skin diseases. Nowadays, there is a considerable development in evolving the transportation system as well, by using advanced driver assistance systems, vehicle detection and much more, thus making driving a safer and more technologically advanced. Other fields include smart homes, object detection in sport videos to be able to detect a ball during the play, rain detection, or even assisting a visually impaired individual.

---

[1] https://coral.ai/docs/edgetpu/faq/

Over recent years, there has been the need to detect and track humans with a relatively high accuracy in real-time. Such task is of a high complexity, because of the swift and sudden movement a human has compared to objects like cars which move in a more predicted trajectory. In order to achieve this goal, various combinations of the aforementioned object detection and tracking methods have been proposed like these presented in [8], [11]. The drawbacks with these methods are that the proposed algorithms involving legacy machine learning are considered obsolete compared to the DL models. This drawback is covered by this paper though the best combinations of DL models and tracking algorithms.

## 3. Background Information

Our detection algorithm is based on two DL models, i.e., the combination between the Single Shot Detector (SSD) and the MobileNet. Our aim is to enhance the detection accuracy being aligned with the needs of the problem of human detection when various changes in the behavior of humans are the case.

**The SSD Model**. The SSD model composes of two parts, i.e., the extraction of feature maps and the application of convolution filters to detect objects. In order to achieve the feature extraction, the model utilises the VGG-16, a CNN used for classification and detection [21]. After the features are extracted, the model detects objects using the Conv4_3 layer, which is a convolution layer being part of the VGG-16. A cross-correlation function is applied to execute four (4) predictions per location or cell and returns a boundary box and 21 scores, resembling the equal amount of classes available. The aforementioned scores represent the probability of each class for this specific boundary box. In addition, the SSD contains 6 auxiliary convolution layers, five of which are added for object detection, and are required in order to support multi-scale feature maps (detection using multiple layers at different scales). The convolutional layers are as follows: (i) Conv4_3 - dimensions: 38X38X4 → 38X38 and 4 detections; (ii) Conv7 - dimensions: 19X19X6→ 19X19 and 6 detections; (iii) Conv8_2 - dimensions: 10X10X6 → 10X10 and 6 detections; (iv) Conv9_2 - dimensions: 5X5X6 → 5X5 and 6 detections; (v) Conv10_2 - dimensions: 3X3X4 → 3X3 and 4 detections; (vi) Conv11_2 - dimensions: 4 → 4 detections. A sum of the provided calculations in each layer will result in 8732 detections per class completed by the model.

**MobileNet Architecture**. The structure of the MobileNet is based on depthwise seperable convolutions in the entirety of its 28 layers, with the exception of the first layer which executes a full convolution. A depthwise seperable convolution, is an operation consisting of two parts; a depthwise convolution, and a pointwise convolution, which are calculated using the respective formulas [12]. All layers are followed by a bachnorm, which is a method to make neural networks faster and more stable by normalizing the input layer, and ReLU (Rectified Linear Unit) activation function. The sole exception of using the ReLU methodology is the final, fully connected layer which has no linearity and uses a softmax layer for classification; useful for the application of cross-entropy loss. The MobileNetV2 uses the same structure as the above, with two main differences; the addition of inverted residuals and linear bottlenecks, i.e., a procedure that produces a number of 53 layers in total. An inverted residual block is a type of residual block that uses an inverted structure for efficiency reasons. The difference between the inverted residual block with its non-inverted counterpart is that first, a pointwise convolution is applied, followed by a 3X3 depthwise convolution, and finishing with a final pointwise convolution in order to be able to add the input and the output, as the number of channels will be reduced.

## 4. Human Detection and Tracking

**The combination of SSD and MobileNet**. To increase the efficiency, we combine the SSD and the MobileNetV2 models as the main bottleneck of the SSD architecture is the VGG-16 backbone. This is because it requires a high computational power. In general, the accuracy of the model is high, however, the efficiency speed-wise is limited. The target is to adopt the discussed model in devices with low computational resources (e.g., mobile devices, IoT nodes) to build efficient pervasive applications. This issue is solved by replacing the SSD's backbone with the MobileNet, resulting in some minor loss in accuracy, but significantly higher efficiency especially in devices without a GPU.

**Detection and Tracking upon SSD & MobileNet**. The optimal combination for human detection and tracking at edge devices is the one using SSD & MobileNetV2 as a detector along with Centroid and Correlation Tracking (CCT) algorithm. The SSD-MobileNetV2 is a lightweight model appropriate for object, therefore, human detection at mobile devices. As for the algorithm utilized for our tracking purposes, the CCT algorithm is, in comparison with other tracking algorithms, lightweight in respect to its spatial demands and suitable for devices with limited hardware capabilities. Centroid tracking is a multi-step process; At first, the detector (the SSD-MobileNetV2) passes in a set of bounding box coordinates for each object detected in every frame. Once received, the center of the coordinates of each of the bounding boxes is calculated. For every new set of coordinates received, thus for every new centroid computed, the Euclidean distance between each pair of existing and new set of centroids has to be calculated. This step is done in order to associate a centroid with another, based on its closest one. If the distance between two centroids, one in the first frame and the other in the second frame, is the smallest compared to all the other ones of the previous frame then a set is selected, meaning that both centroids are describing the same object in these two frames. If a point is left without a pair, a new object will be registered by assigning a new unique ID on it, and storing the centroid of its bounding box as before. Finally, if an old object cannot be matched with a new one, it is deregistered, or else removed from the tracking process. Specifically, in our application domain two parameters are defined to assist with this operation;

maximum disappearance and maximum distance, with the former describing the maximum frames from which a point can exist without being part of a pair and the latter being the maximum distance between two centroids to associate an object. In Figure 1, we can see an example of the proposed approach for the detection and tracking of humans.

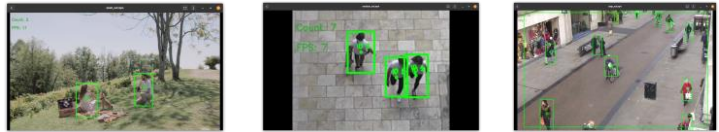

**Figure 1.** An example of the proposed combination of SSD & MobileNetV2.

## 5. Experimental Evaluation

**Experimental Setup**. The proposed model has been trained using the COCO 2014 dataset (https://cocodataset.org) running on an Edge Coral TPU Coral TPUs are appropriate to emulate an edge computing environment where pervasive applications can be executed. The adopted training dataset supports multiple classes-objects and not only humans. However, the dataset is expanded to incorporate more images related to humans for increasing the quality of the process. It is also worth underlining the fact that while SSD-MobileNetV2 could run without any issues or needed changes on edge devices, the Centroid and Correlation Tracking algorithm requires to discard the Scipy dependency, as it is not supported by our TPUs. Instead, we used similar functions from the Numpy dependency in order for it to successfully run on Coral TPUs. We perform our experimentation and detect the performance of the proposed model based on a set of metrics. As the problem under consideration is mainly a classification problem (i.e., detect a human in a aframe and perform the desired tracking), we rely on the widely adopted metrcis Accuracy, Precision and Recall. The following equations hold true: Precision = TP / (TP + FP); Recall = TP / (TP + FN); Accuracy = (TP + TN) / N; where N is the total number of the adopted samples. In the above equations, TP (True Positive) is the number of human detections which have been classified correctly. FP (False Positive) is the number of the absence of humans which have been classified as true detections. FN (False Negatives) is the number of human detections which have not been classified as true detection events. TN (True Negatives) is the number of the absence of human detections that have been correctly classified. The procedure of experimental assessment is performed on three different cases of human gatherings, ranging from low to medium and then high, in order to determine the accuracy, speed and efficiency of the models. The term "low gatherings" means that there is a small amount of people in a video (<5 ). Likewise, "medium gatherings" and "high gatherings" are used to describe the moderate (<10) and big amount of people (>10) in a video, respectively. The aforementioned metrics will be calculated for each of the three videos (named Small, Medium and Large respectively, depending on the number of humans), for nine different thresholds at each one; if the threshold is high, then the minimum acceptable percentage of the model detecting if someone is human or not (in order to create their bounding box) rises as well.

**Experimental Assessment**. Figure 2 depicts how the SSD-MobileNetV2 model reacts in different values of the adopted threshold. By observing them closely, one can come into the conclusion that the model is working exceptionally well for videos with low and medium number of people (from 0 to 10 people). Specifically, for the small video for the threshold value of 0.70, we can obtain the absolute optimal result possible at a detection, reaching a maximum of unity at all three metrics. This means that our detector has all ground truth boxes detected correctly. On the other hand, the model's ability of detecting humans is limited in large crowds, as described in the final plot; especially for values upon 0.70 of the threshold. At that cases, we observe a high precision with a low recall, thus, implying the appearance of many false negatives. A conclusion to draw from this experimental evaluation is that the optimal threshold value that will perform equally well in all situations/videos is 0.50, as it yields high values of precision and recall in all of the three instances, i.e., videos. Last but not least, from small to large videos and from low to high values of threshold respectively, the model's human detecting ability is deteriorating, with a noteworthy zero scored in all three metrics for videos with large crowds and 0.90 threshold value. In any case, the selection of the threshold can be based on the specific problem under consideration paying attention to the outcomes relevant to a specific metric. For instance, we may want to have an increased Precision or Recall depending on the type of the application.

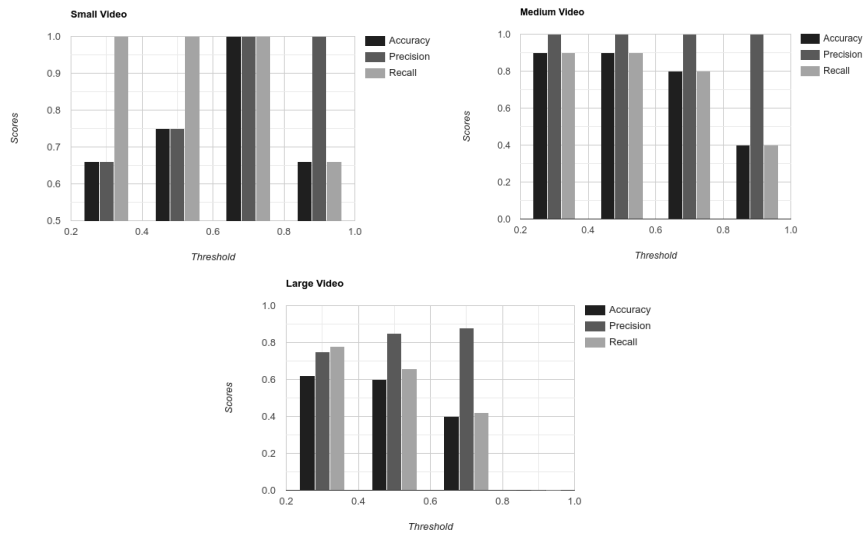

**Figure 2.** Performance outcomes for the SSD-MobileNet-CCT model.

## 6. Conclusions and Future Work

Human detection and tracking upon video streams is an interesting application that can be adopted in various research domains. The target is to detect, count and trace the trajectories of humans while moving in an area under surveillance. This paper proposes a model for human detection and tracing combining and enhancing legacy models to increase the accuracy of the detection. We also argue on their incorporation on embedded devices in order to use them in the edge ecosystem. Our experimental evaluation executed on real datasets reveals the performance of the proposed models and their ability to reach high levels of accuracy. Additionally, we provide the outcomes of our exercise to place these models in embedded devices and describe the relevant requirements. In the first places of our future research agenda is the adoption of forecasting techniques for estimating the future location of humans upon the collected images.